\documentclass{svproc}
\usepackage{url}
\usepackage{doi}
\usepackage{cite}
\usepackage{graphicx}
\usepackage{multirow}
\usepackage{amsmath}
\usepackage{algorithm}%
\usepackage{hhline}%
\usepackage{algorithmicx}%
\usepackage{algpseudocode}%

\begin{document}
\mainmatter              
\title{cuAPO: A CUDA-based Parallelization of Artificial Protozoa Optimizer}           
\titlerunning{cuAPO}  
%
\author{Henish Soliya \and Anugrah Jain}
\authorrunning{Henish Soliya et al.} 
%
\tocauthor{Henish Soliya and Anugrah Jain}
\institute{Sardar Vallabhbhai National Institute of Technology, Surat, Gujarat, India,\\
\email{henish125@gmail.com, ajain@coed.svnit.ac.in}}

\maketitle              

\begin{abstract}
Metaheuristic algorithms are widely used for solving complex problems due to their ability to provide near-optimal solutions. But the execution time of these algorithms increases with the problem size and/or solution space. And, to get more promising results, we have to execute these algorithms for a large number of iterations, requiring a large amount of time and this is one of the main issues found with these algorithms. To handle the same, researchers are now-a-days working on design and development of parallel versions of state-of-the-art metaheuristic optimization algorithms. We, in this paper, present a CUDA-based parallelization of state-of-the-art Artificial Protozoa Optimizer leveraging GPU acceleration. We implement both the existing sequential version and the proposed parallel version of Artificial Protozoa Optimizer for a performance comparison. Our experimental results calculated over a set of CEC2022 benchmark functions demonstrate a significant performance gain i.e. up to 6.7 times speed up is achieved with proposed parallel version. We also use a real world application, i.e., Image Thresholding to compare both algorithms. 
\keywords{Parallel Artificial Protozoa Optimizer, CUDA based Parallelization, High Performance Engineering Optimization}
\end{abstract} 
\section{Introduction}\label{sec1}

Optimization is a fundamental aspect of solving complex real-world problems across the fields. With the rapid increase in complexity of problems in scientific and industrial fields, classical mathematical optimization methods have become ineffective \cite{l2024metaheuristic}. They have several constraints such as early convergence, dependency on derivatives, and having large computing costs particularly for solving NP-hard tasks. These challenges highlighted the need for more robust, efficient and flexible optimization techniques handling multi modal, nonlinear and non differentiable optimization problems effectively \cite{l2024metaheuristic}. In the result, metaheuristic optimization algorithms have emerged and shown their efficiency in solving such problems \cite{l2024metaheuristic,j1995particle,j1992ga,m2005ant,j2023asurvey,p2013backtracking,s2013grey,s2016thewhale,s2016sca,l2021thearithmetic,w2019supply,b2020student,f1998tabu,s1983optimization,k2001aconvergence,h2003iterated,n2022golden,g2017spotted,x2024artificial}.

\par However, metaheuristic optimization algorithms, especially population-based ones, are computationally intensive due to the iterative nature of evaluating and updating a population of solutions . As the dimensionality of problems increases or the evaluation cost of objective functions grows, these algorithms exhibit a significant computational requirement leading to a large amount of execution time \cite{y2023parallel,z2023astudy}. For real-time applications where both speed and quality of solution are important, this becomes a real challenging tasks. To address this issue, performance-enhanced parallel versions of existing optimizers have been developed. By distributing computing jobs over several processors, parallelization have reduced their computation time to a great extent \cite{y2023parallel,z2023astudy}.

\par Artificial Protozoa Optimizer (APO) is the recently proposed bio-inspired metaheuristic algorithm for engineering optimization \cite{x2024artificial}. It is based on the survival behavior of protozoa which includes foraging, dormancy, and reproduction steps for each protozoa in every iteration \cite{x2024artificial}. Foraging step is further classified into autotrophic and heterotrophic foraging. With using autotrophic foraging and dormancy, exploration is performed. Whereas, heterotrophic foraging and reproduction emphasizing on exploitation in solution space \cite{x2024artificial}. This optimizer has already demonstrated its capability by outperforming 32 state-of-the-art algorithms on the challenging CEC2022 benchmark functions in the recent work \cite{x2024artificial}.

\par The APO algorithm is good in terms of getting optimal solutions for large complex problems but its execution time limits its scalability and applicability for real-time tasks. In this paper, we propose a novel CUDA-based parallelization of artificial protozoa optimizer (cuAPO). Under the same, our research contributions made in this paper are as follows.

\begin{itemize}
    \item A detailed discussion to understand state-of-the-art Artificial Protozoa Optimizer is provided. 
    \item A novel CUDA-based parallelization of Artificial Protozoa Optimizer leveraging GPU acceleration for high performance is proposed.   
    \item Both state-of-the-art and proposal parallel version of APO are implemented and compared in terms of performance using well-known CEC2022 benchmark functions. 
    \item A real world application corresponding to Image Thresholding to compare performance of both implementations is also provided.      
\end{itemize}

The rest of the paper is organized as follows. Our next section discusses the state-of-the-art optimizers available in the literature. Thereafter, Section 3 details the existing Artificial Protozoa Optimizer (APO) for readers. In Section 4, we present our proposed parallelization leveraging GPU threads. Our implementation and performance comparison of both sequential and parallel versions of APO are presented in Section 5. Section 6 concludes this paper by presenting some pointers for future work.        

\section{Background}\label{sec2}
Metaheuristic optimizers belong to a powerful class of optimization approaches that can solve a variety of challenging optimization problems due to their flexibility and robust nature. Metaheuristics can handle non-differentiable, multi modal, and nonlinear objective functions and do not rely on derivative information like traditional optimization techniques do \cite{l2024metaheuristic}. When we solve NP-hard problems, achieving accurate solutions are computationally impossible where such metaheuristics seem helpful \cite{l2024metaheuristic}. The metaheuristic optimization algorithms can be classified into the following two categories.     

\subsection{S-Metaheuristics (Single-Solution-Based Algorithms)}
S-metaheuristics are techniques for optimization that only work on a single solution at a time. Compared to their population-based equivalents, these algorithms are easier to construct and more computationally efficient because they iteratively refine their answer by exploring the solution space \cite{l2024metaheuristic}. They work well for problems where local search is more important than global search. These algorithms are more likely to become trapped in local optima due to the absence of population diversity, which may reduce their efficacy for solving complex problems. Tabu Search (TS) \cite{f1998tabu}, Simulated Annealing (SA) \cite{s1983optimization}, Hill Climbing (HC) \cite{k2001aconvergence}, Iterated Local Search (ILS) \cite{h2003iterated} are some of the examples of S-metaheuristics.   

\subsection{P-Metaheuristics (Population-Based Algorithms)}
In this category of metaheuristic algorithms, a population can be defined as the collection of sample solutions generated by the algorithm for consideration in a single iteration. These are also referred to as individuals. These individuals of a population can generate further individuals as offspring with the help of the genetic operators of the procedure \cite{j1992ga}. Because there are many different candidate solutions, it has a high capacity to avoid local optima. It also makes exploration and exploitation easier by allowing candidates to share information. However, it is computationally expensive due to evaluations of multiple solutions in each iteration. Particle Swarm Optimization (PSO) \cite{j1995particle}, Genetic Algorithm (GA) \cite{j1992ga}, Ant Colony Optimization (ACO) \cite{m2005ant}, Backtracking Search Optimization (BSO) \cite{p2013backtracking},  Grey Wolf Optimizer (GWO) \cite{s2013grey}, Whale Optimization Algorithm (WOA) \cite{s2016thewhale}, Sine Cosine Algorithm (SCA) \cite{s2016sca}, Arithmetic Optimization Algorithm (AOA) \cite{l2021thearithmetic}, Supply Demand based Optimization (SDO) \cite{w2019supply}, Student Psychology based Optimization (SPO) \cite{b2020student}, Golden Jackal Optimization (GJO) \cite{n2022golden}, Spotted Hyena Optimizer (SHO) \cite{g2017spotted}, Artificial Protozoa Optimizer (APO) \cite{x2024artificial}, Parallel Particle Swarm Optimization (P-PSO) \cite{y2023parallel}, and Parallel Ant Colony Optimization (P-ACO) \cite{z2023astudy} are some examples of the P-metaheuristic optimization algorithms. \\

Among the aforementioned optimizers, the Artificial Protozoa Optimizer seems the best algorithm and can be taken as the most suitable candidate for proposed parallelization. Based of the various experimental studies conducted in the literature, it has been identified as superior against the existing 32 other state-of-the-art optimization algorithms \cite{x2024artificial}. Hence, we select this for our proposed parallelization so that its computation time can be reduced. Our next section first explains the state-of-the-art Artificial Protozoa Optimizer in detail. 

\section{Artificial Protozoa Optimizer (APO)}\label{sec3}  
\begin{figure}[!ht] 
    \begin{center}
  	\includegraphics[width=\textwidth]{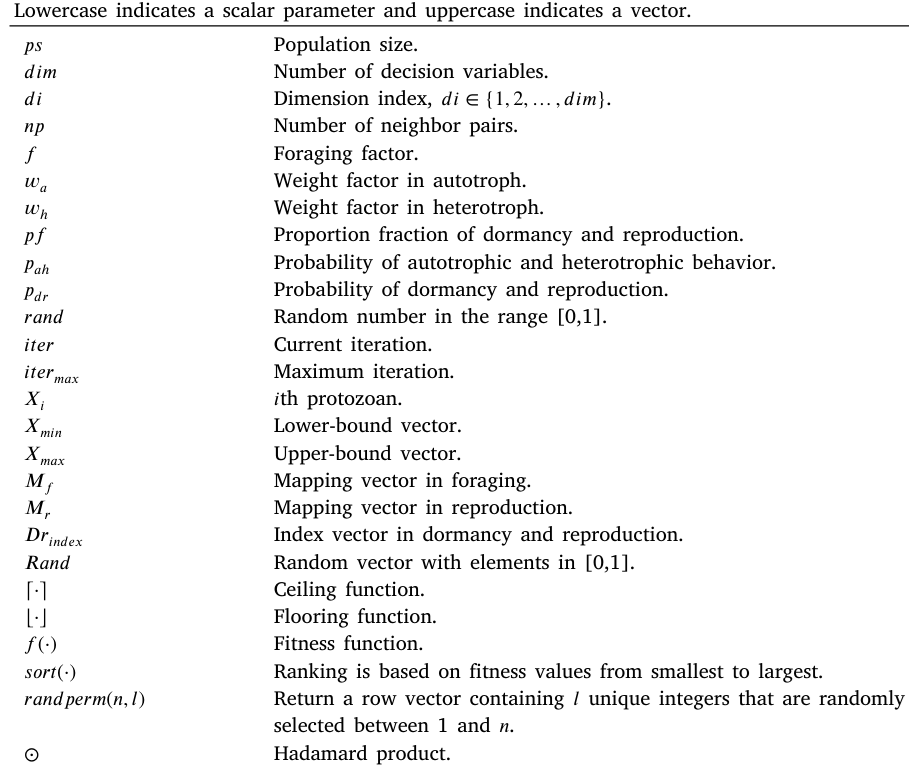}\\
		\caption{List of Notations \cite{x2024artificial}}
		\label{fig:lon}
	\end{center}
\end{figure}

The existing Artificial Protozoa Optimizer (APO) is inspired from the survival strategies of single-celled organism, protozoa \cite{x2024artificial}. Their survival strategies mainly include dormancy, reproduction, foraging in autotrophic mode, and foraging in heterotrophic mode \cite{x2024artificial}. Here, we explain them in detail. Figure \ref{fig:lon} presents a list of notations used by APO. 

\subsection{Dormancy}
During environmental stress, a protozoa choose dormancy as a survival strategy. When the protozoa is dormant, To keep the population constant a newly created protozoa takes its place \cite{x2024artificial}. The following is the mathematical model for dormancy:
\begin{equation}
X_i^{new} = X_{min} + Rand \odot (X_{max} - X_{min})
\label{eq:dormancy}
\end{equation}

\subsection{Reproduction}
Using binary fission, protozoa undergo asexual reproduction. As a result, protozoa divide into two identical daughters. This behavior is simulated by generating a duplicate protozoa. The following is the mathematical model for reproduction:
\begin{equation}
X_i^{new} = X_i \pm rand \cdot \big(X_{min} + Rand \odot (X_{max} - X_{min})\big) \odot M_r
\label{eq:reproduction}
\end{equation}
\begin{equation}
M_r[di] = 
\begin{cases} 
1, & \text{if } di \text{ is in } randperm(dim, \lceil dim \cdot rand \rceil) \\
0, & \text{otherwise}
\end{cases}
\label{eq:mr}
\end{equation}

\subsection{Foraging}
Protozoa has two foraging behavior, i.e., autotrophic mode and heterotrophic mode. The common mathematical model for these two mode is as follows:
\begin{equation}
f = rand \cdot \big(1 + \cos(\frac{iter}{iter_{\max}} \cdot \pi)\big)
\label{eq:f}
\end{equation}
\begin{equation}
M_f[di] = 
\begin{cases} 
1, & \text{if } di \text{ is in } randperm(dim, \lceil dim \cdot \frac{i}{ps} \rceil) \\
0, & \text{otherwise}
\end{cases}
\label{eq:mf}
\end{equation}

\subsubsection{Autotrophic Mode}
Protozoa can synthesize carbohydrates via chloroplasts to take nutrition. The protozoa will shift from its current location to less light intensity location for foraging, when the protozoa exposed to strong light intensity. If the light intensity surrounding the jth protozoa is suitable for the photosynthesis, then the protozoa will move towards the location of the jth protozoa. The following is the mathematical model for autotrophic mode as foraging:
\begin{equation}
X_i^{new} = X_i + f \cdot \big(X_j - X_i + \frac{1}{np} \cdot \sum_{k=1}^{np} w_a \cdot (X_{k-} - X_{k+}) \big) \odot M_f
\label{eq:foragingA}
\end{equation}
\begin{equation}
w_a = e^{-\big|\frac{f(X_{k-})}{f(X_{k+}) + eps}\big|}
\label{eq:wa}
\end{equation}

\subsubsection{Heterotrophic Mode}
A protozoa can get nutrition in the dark environment by consuming organic matter from its nearby food source. If $X_{near}$ is a nearby food source, then the protozoa will move toward it. The following is the mathematical model for heterotrophic mode as foraging:
\begin{equation}
X_i^{new} = X_i + f \cdot \big(X_{near} - X_i + \frac{1}{np} \cdot \sum_{k=1}^{np} w_h \cdot (X_{i-k} - X_{i+k}) \big) \odot M_f
\label{eq:foragingH}
\end{equation}
\begin{equation}
X_{near} = \big(1 \pm Rand \cdot (1 - \frac{iter}{iter_{\max}})\big) \odot X_i
\label{eq:near}
\end{equation}
\begin{equation}
w_h = e^{-\big|\frac{f(X_{i-k})}{f(X_{i+k}) + eps}\big|}
\label{eq:wh}
\end{equation}

\subsection{Working of APO}
Initially, all the protozoa samples are initialized with some random values. Thereafter, their sorting based on fitness value is performed. Each of the samples then update their values based on the above discussed operations. The exploration of the solution space is carried out by autotrophic foraging and dormancy, whereas its exploitation is carried out performing heterotrophic foraging and reproduction \cite{x2024artificial}. This process of updation is repeated for several iterations until it is converged to an optimal solution. Following equations determine the operation selected for every update.   

\begin{equation}
pf = pf_{max} \cdot rand
\end{equation}
\begin{equation}
p_{ah} = \frac{1}{2} \cdot \big(1 + \cos(\frac{iter}{iter_{\max}} \cdot \pi) \big)
\label{eq:pah}
\end{equation}
\begin{equation}
p_{dr} = \frac{1}{2} \cdot \big(1 - \cos((1 - \frac{i}{ps}) \cdot \pi) \big)
\label{eq:pdr}
\end{equation} 

Our next section discusses the proposed CUDA-based parallelization of Artificial Parallel Protozoa Optimizer in detail. An example showing computation of new protozoa values is also presented.     

\begin{figure}[!ht] 
    \begin{center}
  	\includegraphics[width=\textwidth]{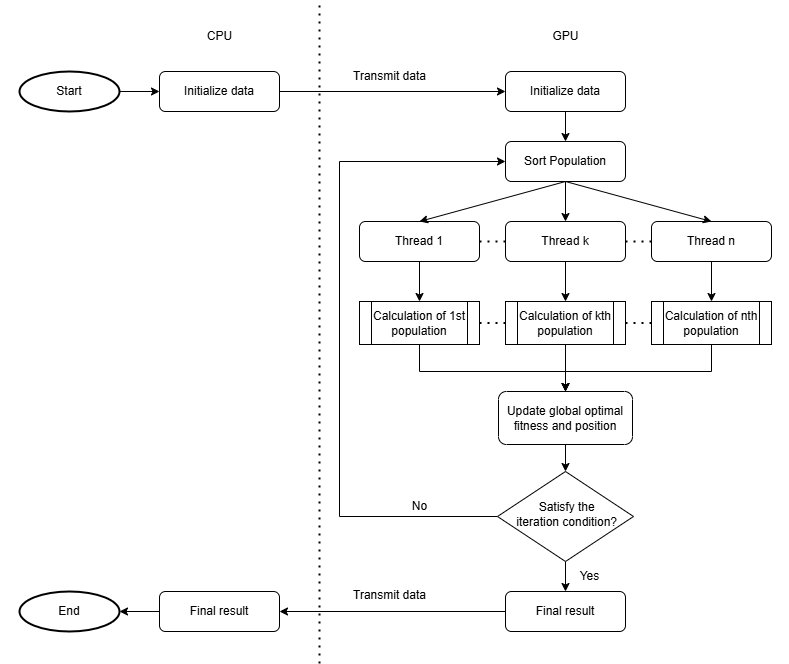}\\
		\caption{Flowchart for cuAPO}
		\label{fig:papo}
	\end{center}
\end{figure}

\section{cuAPO: Proposed CUDA-based Parallelization of Artificial Protozoa Optimizer}\label{sec4}
This section presents our performance-optimized parallelization of state-of-the-art Artificial Protozoa Optimizer. We use CUDA framework for performing parallel computation of protozoa values. A flow chart for proposed parallelization is shown in Figure \ref{fig:papo}.

\begin{algorithm}[!ht] 
\caption{CuAPO: Proposed CUDA-based Parallelization of APO}
\begin{algorithmic}[1]
 
\Statex \textbf{Input}: Set up the parameters np, ps, dim, $pf_{max}$, and MaxFEs.
\Statex \textbf{Output}: Best fitness value $f(X_{gbest})$ and it's position $X_{gbest}$.

\State \textbf{Initialize} parameters on CPU (ps, dim, np, $pf_{max}$, $X_(protozoa)$ and MaxFEs)
\State \textbf{Send} initialized parameters to GPU
\State \textbf{Initialize} data on GPU (threadIDs and assign one protozoa to one thread)

\While{FEs $<$ MaxFEs}
    \State \textbf{Sort} \(X_i\), \(i=1, 2, \ldots, ps\)
    \State \(pf = pf_{max} \cdot rand\)
    \State \(Dr_{index}\) = \(randperm(ps, \lceil ps \cdot pf \rceil )\)

    \State \textbf{Parallel for each} \(i=1:ps\)
        \If{\(i \in Dr_{index}\)}
            \If{\(pdr > rand\)}
                \State Calculate \(X_i^{new}\) using Eq. (\ref{eq:dormancy}) \Comment{dormancy}
            \Else
                \State \(Mr = zeros(1, dim)\)
                \State \(Mr[1, randperm(dim, \lceil dim \cdot rand \rceil)] = 1\)
                \State Calculate \(X_i^{new}\) using Eq. (\ref{eq:reproduction}) \Comment{reproduction}
            \EndIf
        \Else
            \State \(Mf = zeros(1, dim)\)
            \State \(Mf[1, randperm(dim, \lceil dim \cdot \frac{i}{ps} \rceil)] = 1\)
            \If{\(pah > rand\)}
                \State Calculate \(X_i^{new}\) using Eq. (\ref{eq:foragingA}) \Comment{foraging in an autotroph}
            \Else
                \State Calculate \(X_i^{new}\) using Eq. (\ref{eq:foragingH}) \Comment{foraging in a heterotroph}
            \EndIf
        \EndIf

        \If{$f(X_i^{new})$ is optimal then $f(X_i)$}
            \State \(X_i \gets X_i^{new}\)
        \Else
            \State \(X_i \gets X_i\)
        \EndIf
    \State \(FEs \gets FEs + 1\)
\EndWhile

\State \textbf{Send} $X_{gbest}$ and $f(X_{gbest})$ to CPU
\State \textbf{Free} GPU Resources

\end{algorithmic}
\end{algorithm}

At the start, the optimization problem is defined using a mathematical function and same solutions are created in the form of various distinct protozoa values. Each population which is represented by a different protozoa value is updated by a CUDA thread and is happened for each iteration. This process continues until it meets the stopping criteria. Algorithm 1 discusses all steps of cuAPO. Our next subsection discusses an example to illustrate calculation of each protozoa done by a CUDA thread for every iteration.     

\subsection{Calculation of New Protozoa: An Example} 
For the sake of understanding, we only demonstrate the first iteration of optimization process. The same is done for every remaining iterations. To make it simple, Sphere function is used as follows:
\begin{equation}
    Minimize: f(x) = \sum_{i=1}^d x_i^2
    \label{eq:sphere}
\end{equation} 
 
\noindent
The parameters are taken as follows: \\
Population Size (ps) = 4 \\
Dimension (dim) = 3 \\
Xmin = 0 \\
Xmax = 10 \\
Maximum proportion fraction of dormancy and reproduction \(pf_{max}\) = 0.1 \\

\noindent
\textbf{Step 1:}
Initialize the population and calculate it's fitness value using eq. (\ref{eq:sphere}), \\
\(X_1 = [4.8, 7.8, 9.4]\) ; \(f(X_1) = 172.24\) \\
\(X_2 = [3.4, 8.4, 9.0]\) ; \(f(X_2) = 163.12\) \\
\(X_3 = [1.4, 5.0, 8.0]\) ; \(f(X_3) = 90.96\) \\
\(X_4 = [4.5, 1.2, 4.7]\) ; \(f(X_4) = 43.78\) \\

\noindent
\textbf{Step 2:}
Sort protozoa based on it's fitness value, \\
\(f(X_1) = 43.78\) ; \(X_1 = [4.5, 1.2, 4.7]\) \\
\(f(X_2) = 90.96\) ; \(X_2 = [1.4, 5.0, 8.0]\) \\
\(f(X_3) = 163.12\) ; \(X_3 = [3.4, 8.4, 9.0]\) \\
\(f(X_4) = 172.24\) ; \(X_4 = [4.8, 7.8, 9.4]\) \\

\noindent
\textbf{Step 3:}
Calculate index vector for Dormancy and Reproduction. \\
\(Dr_{index}\) = \(randperm(ps, \lceil ps \cdot pf \rceil )\) \\
\(Dr_{index}\) = [2] \\

\noindent
\textbf{Step 4:}
Based on the value of \(Dr_{index}\), the protozoa 2 is selected for performing Dormancy/Reproduction. The final operation from these two choices is again selected from equation \ref{eq:pdr}. And, the operations for remaining protozoa are selected using equation \ref{eq:pah}. Finally, the identified operations for all of them are as follows. 

\begin{itemize}
    \item Protozoa 1 is doing Foraging in Autotroph Mode. 
    \item Protozoa 2 is doing Reproduction.
    \item Protozoa 3 is doing Foraging in Heterotroph Mode.
    \item Protozoa 4 is doing Foraging in Autotroph Mode.
\end{itemize}

\noindent
\textbf{Step 5:}
The new values for each protozoa after performing the aforementioned operations are as follows. 

\begin{itemize}
    \item \(X_1^{new} = [4.5, 1.77, 4.7]\) 
    \item \(X_2^{new} = [1.4, 13.53, 11.53]\)
    \item \(X_3^{new} = [2.9, 8.38, 9.11]\)
    \item \(X_4^{new} = [4.26, 6.32, 8.07]\)
\end{itemize}

\noindent
\textbf{Step 6:}
This step finalize, the values of all protozoa for next iteration. First the above values are kept in the earlier mentioned range. Afterwards, the selection between the old and new values of protozoa is performed as follows. \\

\noindent
\begin{tabular}{ |l|l| }
    \hline
    \textbf{Old Position} & \textbf{New Position} \\
    \hline

    \(f(X_1) = 43.78\) ; \(X_1 = [4.5, 1.2, 4.7]\) & \(f(X_1^{new}) = 45.47\) ; \(X_1^{new} = [4.5, 1.77, 4.7]\) \\
    
    \(f(X_2) = 90.96\) ; \(X_2 = [1.4, 5.0, 8.0]\) & \(f(X_2^{new}) = 201.96\) ; \(X_2^{new} = [1.4, 10.0, 10.0]\) \\
    
    \(f(X_3) = 163.12\) ; \(X_3 = [3.4, 8.4, 9.0]\) & \(f(X_3^{new}) = 161.63\) ; \(X_3^{new} = [2.9, 8.38, 9.11]\) \\
    
    \(f(X_4) = 172.24\) ; \(X_4 = [4.8, 7.8, 9.4]\) & \(f(X_4^{new}) = 123.22\) ; \(X_4^{new} = [4.26, 6.32, 8.07]\) \\

    \hline         
\end{tabular} \\

\noindent
Following are the final values of protozoa for the next iteration.  \\
\(X_1 = [4.5, 1.2, 4.7]\) \\
\(X_2 = [1.4, 5.0, 8.0]\) \\
\(X_3 = [2.9, 8.38, 9.11]\) \\
\(X_4 = [4.26, 6.32, 8.07]\) \\ 

\begin{table}[!ht]
    \caption{Hardware, Software and GPU Kernel Specifications}
    \begin{tabular}{ |l|p{51mm}| }
    
        \hline
        \textbf{Hardware Specifications} & \textbf{Software Specifications} \\
        \hline
        
        \textbf{GPU}: NVIDIA Tesla T4 & \textbf{OS}: Ubuntu 22.04.3 LTS \\
        
        \textbf{GPU RAM}: 15.0 GB & \textbf{GPU Driver Version}: 535.104.05 \\
        
        \textbf{System RAM}: 12.7 GB & \textbf{CUDA Version}: 12.2 \\
        
        \textbf{Disk}: 112.6 GB & \textbf{nvcc Version}: 12.2 \\
        
        \textbf{CPU}: Intel(R) Xeon(R) CPU @ 2.20GHz & \textbf{GCC Version}: 11.4 \\

          & \textbf{Programming Language}: CUDA C++ \\
        \hline
         
    \end{tabular}  

    \begin{tabular}{|p{112mm}|}
    
        \hline
        \textbf{GPU Kernel Specifications: For Population Size (PS) = 10,000} \\
        \hline
        Threads per Block: 500 \\
        Number of CUDA Blocks: 10,000/500 = 20 \\
        Warps per Block: 500 / 32 = 16 \\
        Total Warps: 20 x 16 = 320 \\
        
        \hline
         
    \end{tabular}
    \label{tab:spec}
\end{table}
 
\begin{table}[!ht]
    \centering
    \caption{CEC2022 Benchmark Results}
    \begin{tabular}{|p{0.5cm}|p{2cm}|p{0.5cm}|p{1.5cm}|p{1.5cm}|p{1.5cm}|p{1.5cm}|p{1.5cm}|}
    \hline
    \multicolumn{8}{|c|}{\textbf{Dim = 1000, Iter = 1000, Range = [-100,100]}} \\
    \hline
    \textbf{No.} & \textbf{Function Name} & \textbf{PS (k)} & \multicolumn{2}{c|}{\textbf{Sequential Algorithm}} & \multicolumn{2}{c|}{\textbf{Parallel Algorithm}} & \textbf{Speedup} \\
    \cline{4-7}
     & & & \textbf{Avg. Best Fit} & \textbf{Avg. Time (s)} & \textbf{Avg. Best Fit} & \textbf{Avg. Time (s)} & \\
    \hline
    1 & Bent Cigar & 1 & 9.01E+08 & 211 & 7.95E+08 & 35 & 6.03 \\
      &                     & 2 & 3.24E+08 & 423 & 3.51E+08 & 64 & 6.61 \\
      &                     & 5 & 1.01E+08 & 1040 & 1.03E+08 & 158 & 6.58 \\
      &                     & 10 & 5.06E+07 & 2101 & 5.40E+07 & 345 & 6.09 \\
    \hline
    2 & High Conditioned Elliptic & 1 & 2.33E+06 & 238 & 2.17E+06 & 67 & 3.55 \\
      &                                   & 2 & 806290 & 478 & 761592 & 129 & 3.71 \\
      &                                   & 5 & 346675 & 1180 & 326084 & 322 & 3.66 \\
      &                                   & 10 & 148573 & 2385 & 156662 & 673 & 3.54 \\
    \hline
    3 & HGBat & 1 & 2.19E+07 & 212 & 2.12E+07 & 34 & 6.24 \\
      &                & 2 & 1.47E+07 & 430 & 1.43E+07 & 64 & 6.72 \\
      &                & 5 & 1.05E+07 & 1059 & 1.15E+07 & 160 & 6.62 \\
      &                & 10 & 1.09E+07 & 2135 & 1.05E+07 & 345 & 6.19 \\
    \hline
    4 & Rosenbrock’s & 1 & 1.21E+06 & 285 & 968158 & 96 & 2.97 \\
      &                       & 2 & 255588 & 576 & 241539 & 190 & 3.03 \\
      &                       & 5 & 53900 & 1408 & 40864.2 & 471 & 2.99 \\
      &                       & 10 & 16418 & 2839 & 17000 & 971 & 2.92 \\
    \hline
    5 & Griewank’s & 1 & 0.924789 & 238 & 0.907354 & 48 & 4.96 \\
      &                     & 2 & 0.476121 & 477 & 0.450114 & 93 & 5.13 \\
      &                     & 5 & 0.189392 & 1164 & 0.218021 & 232 & 5.02 \\
      &                     & 10 & 0.094324 & 2348 & 0.096002 & 492 & 4.77 \\
    \hline
    \end{tabular}
    \label{tab:benchmarkresults}
\end{table}

\section{Implementation and Results}\label{sec5}
This section discusses our implementation of both sequential and proposed parallel versions of Artificial Protozoa Optimizer. Its hardware, software and GPU kernel specifications are mentioned in Table \ref{tab:spec}. We use google colab Tesla T4 GPU for achieving parallelization. As shown in the table, our proposed parallelization sets the total number of threads per block as 500 and the total number of blocks is based on the ratio of the total population size and the number of threads per block. Our results achieved for CEC2022 benchmarks functions are discussed next.  

\subsection{Performance on Benchmark Functions}
To compare the both version of APO, we use five CEC2022 benchmark functions, i.e., Bent Cigar Function, High Conditioned Elliptic Function, HGBat Function, Rosenbrock’s Function, and Griewank’s Function. Table \ref{tab:benchmarkresults} shows our performance evaluation of both sequential and parallel APO algorithms on the mentioned CEC2022 benchmark functions. Each result is averaged over the five simulation runs. The performance evaluation shown here clearly states that the proposed parallelization is performance efficient, reduces running time significantly. The achieved average speedup is 4.87, with the best speedup is 6.72 and the worst speedup is 2.92.  

\begin{figure}[H]  
    \begin{center}
  	\includegraphics[width=\textwidth]{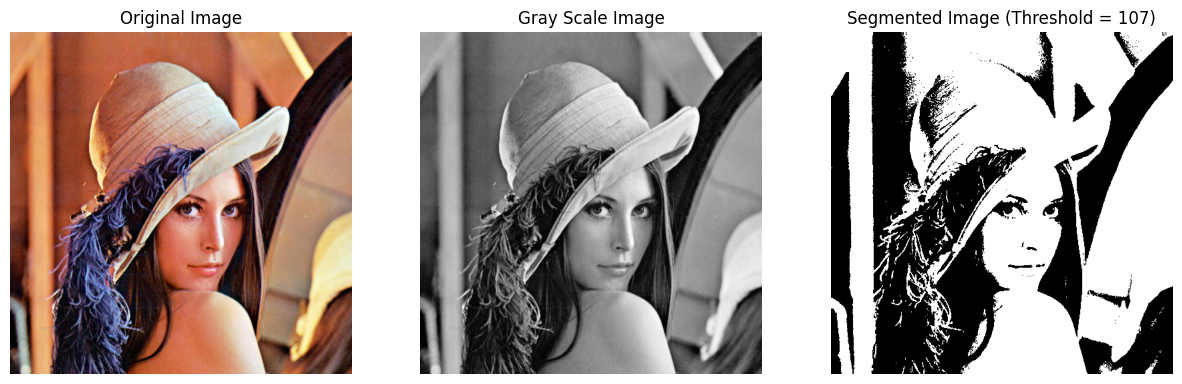}
		\caption{Image Thresholding Example}
		\label{fig:lena}
	\end{center}
\end{figure}      
 
\subsection{Image Thresholding: A Real Application}
This subsection applies both versions for image thresholding. In this application, for an input colored image, we find an optimal threshold value for converting it into a purely black and white image, as shown in Figure \ref{fig:lena}. The method of conversion is based on maximizing the between-class variance of the given image for finding the best threshold, popularly known as Otsu's method. To conduct this experiment, four image sizes, 100 population size, and 50 iterations are used. It has been observed that the optimal value of threshold can be identified after performing 50 iterations only. Table \ref{tab:lenaresults} shown our performance results for image thresholding application where each result is averaged over 5 simulation runs. From this experiment, it can be stated again that the proposed parallelization is performance efficient, with an average speedup of 2.2 for the mentioned real application.      

\begin{table}[!ht]
    \centering
    \caption{Results for Image Thresholding}
    \begin{tabular}{|p{2.0cm}|p{1.5cm}|p{1.5cm}|p{1.5cm}|p{1.5cm}|c|}
    \hline
    
    \multicolumn{6}{|c|}{\textbf{PS = 100, Dim = 1, Iter = 50, Range = [0,255]}} \\
    \hline
    
    \textbf{Size (Px)} & \multicolumn{2}{c|}{\textbf{Sequential Algorithm}} & \multicolumn{2}{c|}{\textbf{Parallel Algorithm}} & \textbf{Speedup} \\
    \hhline{|~|-|-|-|-|~|}
    
    & \textbf{Avg. Best Th.} & \textbf{Avg. Exe. Time (s)} & \textbf{Avg. Best Th.} & \textbf{Avg. Exe. Time (s)} & \\ 
    \hline
        
    512 × 512 & 107.17 & 8 & 107.33 & 3 & 2.67 \\
    \hline

    1024 × 1024 & 104.15 & 26 & 104.45 & 13 & 2.00 \\
    \hline

    2048 × 2048 & 104.20 & 103 & 104.56 & 50 & 2.06 \\
    \hline

    4096 × 4096 & 101.43 & 404 & 101.56 & 199 & 2.03 \\
    \hline

    \end{tabular} 
    \label{tab:lenaresults}
\end{table}

\section{Conclusion}\label{sec7}
In this paper, we proposed a novel CUDA-based parallelization of bio-inspired Artificial Protozoa Optimizer (cuAPO), leveraging GPU threads for high performance.  Artificial protozoa optimizer proposed as a better optimization algorithm in literature but with a considerable performance issue. We have reduced its computation time by applying parallel computing using GPU threads. The existing sequential APO was also implemented for the performance comparison. We have used 5 different CEC2022 benchmark functions and a real application (Image Thresholding) for the performance evaluation. The average speedup achieved on CEC2022 benchmark functions is 4.87, whereas for the real application, it is 2.67, respectively. In the near future, we can explore more such parallel version of recently proposed metaheuristic optimizers and apply them for solving various time-bound real-world optimization problems various high performance is needed.


\begin{thebibliography}{23}

\bibitem{l2024metaheuristic}
L. Abualigah, \textit{Metaheuristic Optimization Algorithms}. Elsevier, Amsterdam, Netherlands (2024). \doi{10.1016/C2022-0-02901-9}

\bibitem{j1995particle}
J. Kennedy, R. Eberhart, Particle swarm optimization. In: \textit{Proc. IEEE Int. Conf. Neural Networks (ICNN)}, IEEE (1995). \doi{10.1109/ICNN.1995.488968}

\bibitem{j1992ga}
J.H. Holland, Genetic Algorithms. \textit{Scientific American}, \textbf{267}, 66--72 (1992). \doi{10.1038/scientificamerican0792-66}

\bibitem{m2005ant}
M. Dorigoa, C. Blumb, Ant colony optimization theory: A survey. \textit{Theoretical Computer Science}, \textbf{344}, 243--278 (2005). \doi{10.1016/j.tcs.2005.05.020}

\bibitem{j2023asurvey}
J.-S. Pan, P. Hu, V. Snášel, S.-C. Chu, A survey on binary metaheuristic algorithms and their engineering applications. \textit{Artificial Intelligence Review}, \textbf{56}, 6101--6167 (2023). \doi{10.1007/s10462-022-10328-9}

\bibitem{p2013backtracking}
P. Civicioglu, Backtracking Search Optimization Algorithm for numerical optimization problems. \textit{Appl. Math. Comput.}, \textbf{219}, 8121--8144 (2013). \doi{10.1016/j.amc.2013.02.017}

\bibitem{s2013grey}
S. Mirjalili, S.M. Mirjalili, A. Lewis, Grey Wolf Optimizer. \textit{Adv. Eng. Softw.}, \textbf{69}, 46--61 (2013). \doi{10.1016/j.advengsoft.2013.12.007}

\bibitem{s2016thewhale}
S. Mirjalili, A. Lewis, The whale optimization algorithm. \textit{Adv. Eng. Softw.}, \textbf{95}, 51--67 (2016). \doi{10.1016/j.advengsoft.2016.01.008}

\bibitem{s2016sca}
S. Mirjalili, SCA: A sine cosine algorithm for solving optimization problems. \textit{Knowledge-Based Systems}, \textbf{96}, 120--133 (2016). \doi{10.1016/j.knosys.2015.12.022}

\bibitem{l2021thearithmetic}
L. Abualigah, A. Diabat, S. Mirjalili, M. Abd Elaziz, A.H. Gandomi, The arithmetic optimization algorithm. \textit{Comput. Methods Appl. Mech. Engrg.}, \textbf{376}, 113609 (2021). \doi{10.1016/j.cma.2020.113609}

\bibitem{w2019supply}
W. Zhao, L. Wang, Z. Zhang, Supply-demand-based optimization: A novel economics-inspired algorithm for global optimization. \textit{IEEE Access}, \textbf{7}, 73182--73206 (2019). \doi{10.1109/ACCESS.2019.2918753}

\bibitem{b2020student}
B. Das, V. Mukherjee, D. Das, Student psychology based optimization algorithm: A new population-based optimization algorithm for solving optimization problems. \textit{Adv. Eng. Softw.}, \textbf{146}, 102804 (2020). \doi{10.1016/j.advengsoft.2020.102804}

\bibitem{f1998tabu}
F. Glover, M. Laguna, \textit{Tabu Search}. Springer, Boston, MA (1998). \doi{10.1007/978-1-4615-6089-0}

\bibitem{s1983optimization}
S. Kirkpatrick, C.D. Gelatt Jr., M.P. Vecchi, Optimization by simulated annealing. \textit{Science}, \textbf{220}(4598), 671--680 (1983). \doi{10.1126/science.220.4598.671}

\bibitem{k2001aconvergence}
K.A. Sullivan, S.H. Jacobson, A convergence analysis of generalized hill climbing algorithms. \textit{IEEE Trans. Autom. Control}, \textbf{46}(8), 1288--1293 (2001). \doi{10.1109/9.940936}

\bibitem{h2003iterated}
H.R. Lourenço, O.C. Martin, T. Stützle, \textit{Iterated Local Search}. Springer, Boston, MA (2003). \doi{10.1007/0-306-48056-5\_11}

\bibitem{n2022golden}
N. Chopra, M.M. Ansari, Golden jackal optimization: A novel nature-inspired optimizer for engineering applications. \textit{Expert Systems with Applications}, \textbf{198}, 116924 (2022). \doi{10.1016/j.eswa.2022.116924}

\bibitem{g2017spotted}
G. Dhiman, V. Kumar, Spotted hyena optimizer: A novel bio-inspired metaheuristic technique for engineering applications. \textit{Adv. Eng. Softw.}, \textbf{114}, 48--70 (2017). \doi{10.1016/j.advengsoft.2017.05.014}

\bibitem{x2024artificial}
X. Wanga, V. Snášela, S. Mirjalilib, J. Panc, L. Konga, H.A. Shehadehd, Artificial Protozoa Optimizer (APO): A novel bio-inspired metaheuristic algorithm for engineering optimization. \textit{Knowledge-Based Systems}, \textbf{295}, 111737 (2024). \doi{10.1016/j.knosys.2024.111737}

\bibitem{y2023parallel}
Y. Zhuo, T. Zhang, F. Du, R. Liu, A parallel particle swarm optimization algorithm based on GPU/CUDA. \textit{Appl. Soft Comput.}, \textbf{144}, 110499 (2023). \doi{10.1016/j.asoc.2023.110499}

\bibitem{z2023astudy}
Z. Zeng, Y. Cai, Z. Wu, H. Wang, K.L. Chung, A Study of Parallel Ant Colony Optimization with CUDA-Based Implementation. In: \textit{Proc. ICEICT}, IEEE (2023). \doi{10.1109/ICEICT57916.2023.10245802}

\bibitem{a2023tension}
A. Tzanetos, M. Blondin, A qualitative systematic review of metaheuristics applied to tension/compression spring design problem: Current situation, recommendations, and research direction. \textit{Eng. Appl. Artif. Intell.}, \textbf{118}, 105521 (2023). \doi{10.1016/j.engappai.2022.105521}

\bibitem{k2021problem}
A. Kumar, K. PV, A.W. Mohamed, A.A. Hadi, P. Suganthan, Problem definitions and evaluation criteria for the CEC 2022 special session and competition on single objective real-parameter numerical optimization. Nanyang Technological University, Singapore (2021).

\end{thebibliography}
\end{document}